MEDSAM-BASED LUNG MASKING FOR MULTI-LABEL CHEST X-RAY CLASSIFICATION

Brayden Miao[1], Zain Rehman[1], Xin Miao[2]*, Siming Liu[3], Jianjie Wang[4]

[1]Central High School, Springfield, MO, United States of America
[2]School of Earth, Environment and Sustainability, Missouri State University, Springfield, MO 65897, United States of America
[3]Computer Science Department, Missouri State University, Springfield, MO 65897, United States of America
[4]School of Health Sciences, Missouri State University, Springfield, MO 65897, United States of America

December 28, 2025

## Abstract

Chest X-ray (CXR) imaging is widely used for screening and diagnosing pulmonary abnormalities, yet automated interpretation remains challenging due to weak disease signals, dataset bias, and limited spatial supervision. Foundation models for medical image segmentation (MedSAM) provide an opportunity to introduce anatomically grounded priors that may improve robustness and interpretability in CXR analysis. We propose a segmentation-guided CXR classification pipeline that integrates MedSAM as a lung region extraction module prior to multi-label abnormality classification. MedSAM is fine-tuned using a public image-mask dataset from Airlangga University Hospital. We then apply it to a curated subset of the public NIH CXR dataset to train and evaluate deep convolutional neural networks for multi-label prediction of five abnormalities (Mass, Nodule, Pneumonia, Edema, and Fibrosis), with the normal case (No Finding) evaluated via a derived score. Experiments show that MedSAM produces anatomically plausible lung masks across diverse imaging conditions. We find that masking effects are both task-dependent and architecture-dependent. ResNet50 trained on original images achieves the strongest overall abnormality discrimination, while loose lung masking yields comparable macro AUROC but significantly improves No Finding discrimination, indicating a trade-off between abnormality-specific classification and normal case screening. Tight masking consistently reduces abnormality level performance but improves training efficiency. Loose masking partially mitigates this degradation by preserving perihilar and peripheral context. These results suggest that lung masking should be treated as a controllable spatial prior selected to match the backbone and clinical objective, rather than applied uniformly.

---

* Corresponding author E-mail: xinmiao@missouristate.edu

# 1. Introduction

Chest X-ray (CXR) remains one of the most widely used imaging modalities for screening, triage, and longitudinal monitoring of pulmonary disease because it is fast, inexpensive, and available in most clinical settings [1, 2]. However, CXR interpretation is inherently challenging: many pathologies present with subtle opacities, overlapping anatomical structures, and confounding acquisition variability. These factors contribute to diagnostic uncertainty and motivate the development of computer-aided diagnosis tools that can assist clinicians [2, 3].

Over the past decade, deep learning has become the dominant approach for automated CXR diagnosis, largely enabled by hospital-scale datasets labeled from radiology reports and the rise of transfer learning with high-capacity convolutional neural networks (CNN) [3, 4]. Many methods treat the task as end-to-end image classification, mapping a full CXR image directly to one or more disease labels. While these models can achieve strong benchmark performance, the literature has repeatedly emphasized two persistent weaknesses: weak supervision (labels are image-level and often noisy) and limited generalization across institutions, scanners, and patient populations [3, 5, 6]. As a result, models may learn dataset-specific shortcuts or exploit spurious correlations unrelated to true pathology, which undermines clinical reliability.

A long-standing strategy to improve robustness and interpretability is to explicitly constrain analysis to anatomically relevant regions. Lung field segmentation is a particularly common preprocessing step in CXR pipelines because it reduces background artifacts (e.g., borders, markers) and encourages downstream models to focus on pulmonary parenchyma [6, 7]. Early segmentation approaches relied on classical image processing and machine learning methods, including thresholding, region-based and edge-based techniques, pattern recognition, and deformable models, which often required substantial manual tuning and intervention [8]. In the past decade, deep learning architectures such as U-Net, ResNet, and their variants have been widely adopted for medical image segmentation [7-12]. However, these task-specific segmentation networks were trained on limited mask-annotated data, which can restrict portability across datasets and clinical environments. Recent progress in segmentation foundation models offers a new paradigm. MedSAM adapts the Segment Anything Model (SAM) to medical imaging by fine-tuning a promptable model on a large-scale, multi-modal dataset of medical image–mask pairs, showing strong robustness across diverse segmentation tasks [13]. However, its utility as a portable lung region extractor within CXR diagnosis pipelines has not been systematically characterized.

The evolution of the CXR image segmentation and classification pipeline is summarized in Figure 1. The figure outlines three major research paradigms: traditional machine learning, deep learning, and AI foundation models, applied either to a segmentation-classification pipeline or to end-to-end classification. The AI foundation model (MedSAM) is connected with dashed lines to indicate a research gap. It also presents a time-series analysis of 1,940 articles published over the past 28 years retrieved from the Web of Science in December 2025 using keyword searches for "chest X-ray" and "image segmentation" and/or "classification". Notably, there is a sharp increase in publication volume from 2020-2022, coinciding with the global COVID-19 pandemic.

In this paper, we investigate the use of the foundation model MedSAM as a lung segmentation module within a segmentation-guided CXR diagnosis pipeline. Our contributions are threefold: (1) we fine-tune MedSAM using a public CXR lung mask dataset and apply it to generate reliable lung masks on NIH chest radiographs; (2) we quantify the impact of MedSAM-based segmentation on downstream multi-label abnormality classification across two backbones (DenseNet121 and ResNet50) and different masking strategies; and (3) we provide quantitative and qualitative analyses that clarify the strengths and limitations of MedSAM-based segmentation for CXR diagnosis.

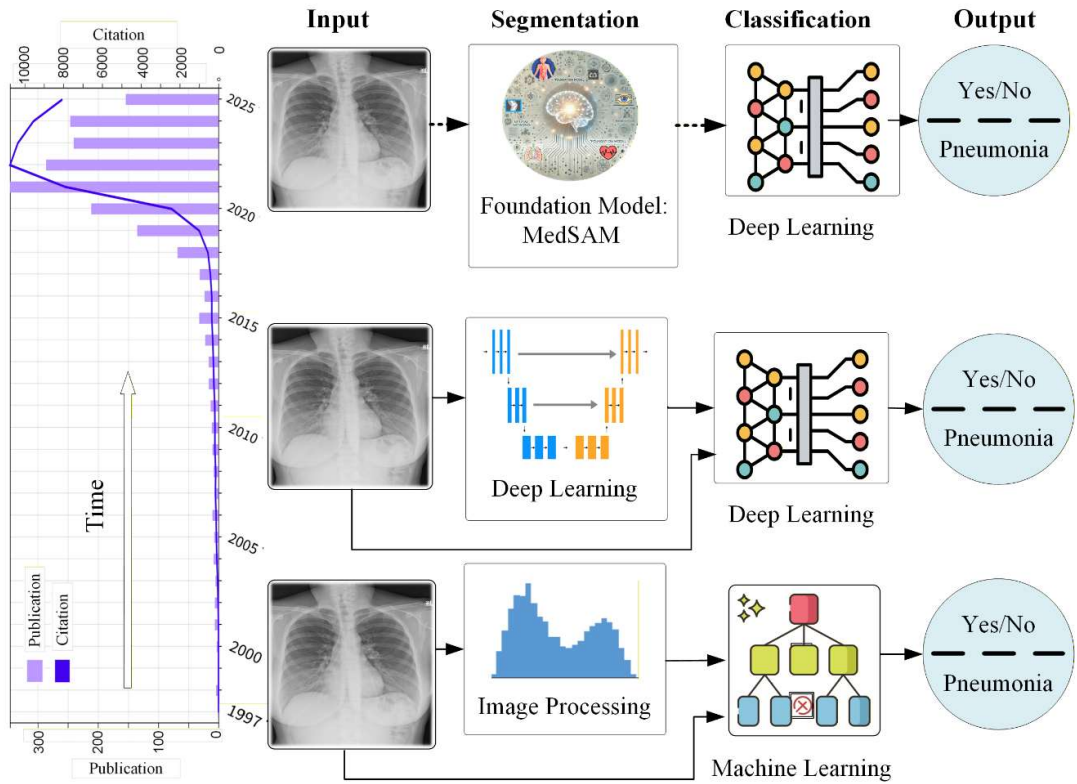

Figure 1. Conceptual illustration of the evolution of CXR segmentation and classification, based on a comprehensive literature review. Publication volume rises sharply from 2020 to 2022, coinciding with the global COVID-19 pandemic.

## 2. Related work

### 2.1. Lung segmentation in chest radiography

Early lung segmentation methods in chest radiography relied on classical image processing and handcrafted features, but these approaches were often brittle under changes in acquisition conditions, pathology severity, or anatomical variation [8]. Deep learning substantially improved segmentation performance and usability. U-Net, in particular, introduced a general encoder–decoder architecture with skip connections that became the backbone of many medical segmentation systems and catalyzed broad adoption across imaging tasks [14, 15]. In the CXR domain, U-Net and ResNet variants and specialized architectures have been proposed to improve boundary delineation, handle intensity inhomogeneity, and increase robustness to artifacts [16, 17]. Complementary work has shown that preprocessing (e.g., normalization and contrast enhancement) can further improve segmentation stability [11], while newer designs such as FCA-Net report accuracy improvements for lung region segmentation on chest radiographs [18].

Foundation models for segmentation seek to overcome the narrow scope of task-specific networks by learning generalizable representations that transfer across tasks and modalities. The Segment Anything Model (SAM), introduced by Meta AI in April 2023 [19], marks a significant milestone as the first open source foundation model for image segmentation. Trained on Meta's massive SA-1B dataset, which consists of 11 million images and 1.1 billion segmentation masks, SAM delivers strong zero-shot performance across a wide array of segmentation tasks. Built on SAM, MedSAM is a medical domain adaptation of a foundation segmentation model trained on large scale medical data [13]. MedSAM retains SAM's core components, including an image encoder, prompt encoder, and mask decoder, but it is fine-tuned on a large-scale medical segmentation corpus to improve robustness on targets with weak boundaries or low contrast. In particular, MedSAM was trained on 1,570,263 image mask pairs spanning 10 imaging modalities and more than 30 cancer types, supporting generalization across heterogeneous clinical imaging conditions [13]. This work positions MedSAM as a general-purpose segmentation component for CXR diagnosis pipelines, with emphasis on fine-tuning MedSAM and quantifying downstream classification benefits, and identifying practical limitations.

### 2.2. CXR classification with large-scale datasets and transfer learning

Progress in CXR abnormality classification has been strongly tied to the release of large-scale datasets labeled from radiology reports, including hospital-scale collections that enable multi-label learning and weakly supervised localization [4, 20]. Using such datasets, deep learning models such as DenseNet, ResNet, and their variants trained with transfer learning have produced competitive results for thoracic disease labeling [1, 3]. However, faithful comparison across studies remains difficult due to differences in preprocessing, dataset splits, and evaluation protocols, and reported performance can vary significantly depending on how datasets are partitioned [3]. More broadly, surveys and systematic reviews emphasize that weak labels, dataset bias, and domain shift continue to limit model generalization and clinical translation [2, 6]. Toolkits such as TorchXRayVision have been developed to improve reproducibility by standardizing access to multiple datasets and providing common preprocessing and baseline models, supporting more consistent evaluation and study of dataset shift [5].

### 2.3. Segmentation–classification pipelines

To improve robustness and interpretability, segmentation–classification pipelines have been proposed to extract lung regions or other anatomical regions of interest (ROIs) before training classifiers. This design aims to reduce reliance on irrelevant image regions and encourage disease prediction from pulmonary anatomy. Empirical studies in X-ray diagnosis, including COVID-era pipelines, report that segmentation modules can improve classification performance and clarify model behavior, though results depend on mask quality and dataset composition [7, 21]. Systematic reviews similarly note that while segmentation-guided workflows are promising, most approaches remain task- and dataset-specific, limiting generalization across institutions and acquisition protocols [6].

## 3. Data and methods

Our proposed CXR segmentation–classification pipeline is illustrated in Figure 2. First, the mask dataset derived from Airlangga University Hospital (RSUA) is used to fine-tune MedSAM, enabling the model to adapt specifically to lung mask segmentation in chest X-ray images [10]. Next, image morphological operations are applied to expand the lung masks, incorporating adjacent thoracic structures for broader anatomical coverage. In the third step, the fine-tuned MedSAM is deployed on a selected NIH CXR dataset, targeting five commonly observed thoracic conditions—mass, nodule, pneumonia, edema, and fibrosis—along with normal cases. Finally, the masked NIH CXR images are fed into deep learning neural networks to evaluate classification performance.

### 3.1. Segmentation and classification datasets

Publicly available CXR datasets were used for segmentation and classification experiments. The CXR segmentation dataset collected from RSUA was used for fine-tuning MedSAM [10]. The RSUA dataset consists of 292 CXR images and their corresponding ground truth (lung masks), which were initially generated by a U-Net-based segmentation model and subsequently validated by radiologists. The dataset includes 207 images from COVID-19 patients, 53 images from pneumonia patients, and 32 images from normal cases. Both the images and masks are provided in BMP format with a resolution of 256×256 pixels. The 292 image-mask pairs were split into three datasets: 70% for fine-tuning MedSAM, 15% for validation, and 10% for testing the performance of lung segmentation.

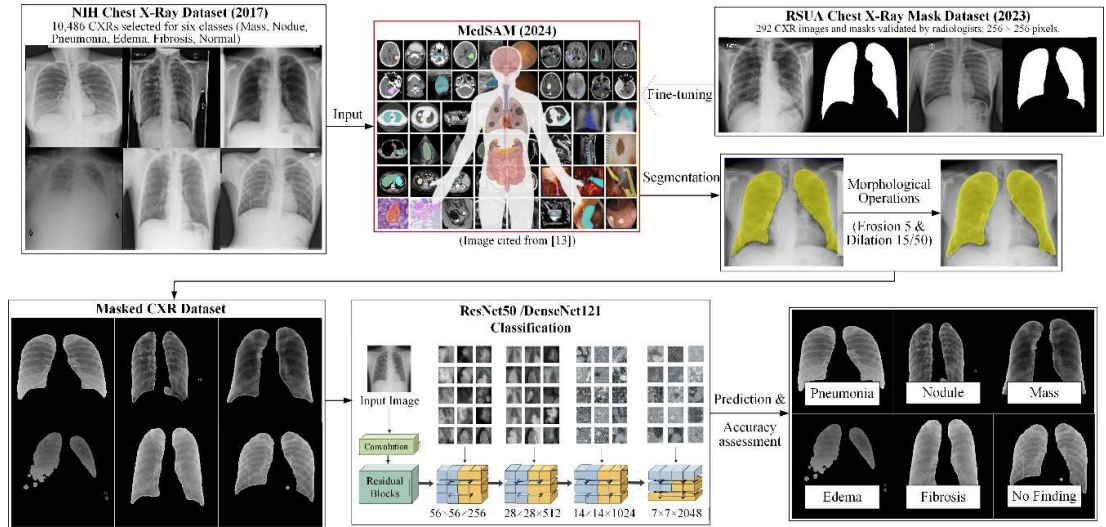

Figure 2. Workflow of the CXR segmentation–classification pipeline.

CXRs and pulmonary abnormality labels for classification were obtained from the NIH CXR dataset (ChestX-ray14), a large-scale repository comprising 112,120 frontal CXR images with 14 abnormality labels from 30,805 unique patients [3, 4]. Each image is provided at a resolution of 512 × 512 pixels, and 14 abnormality condition labels and normal label were obtained via NLP mining of radiology reports, and overall precision, recall, and F1-score were reported of approximately 0.90 [4]. However, as illustrated in Figure 3, the dataset exhibits substantial class imbalance across conditions, with the No Finding (normal) label accounting for 53.8% of samples, which poses challenges for model training and evaluation.

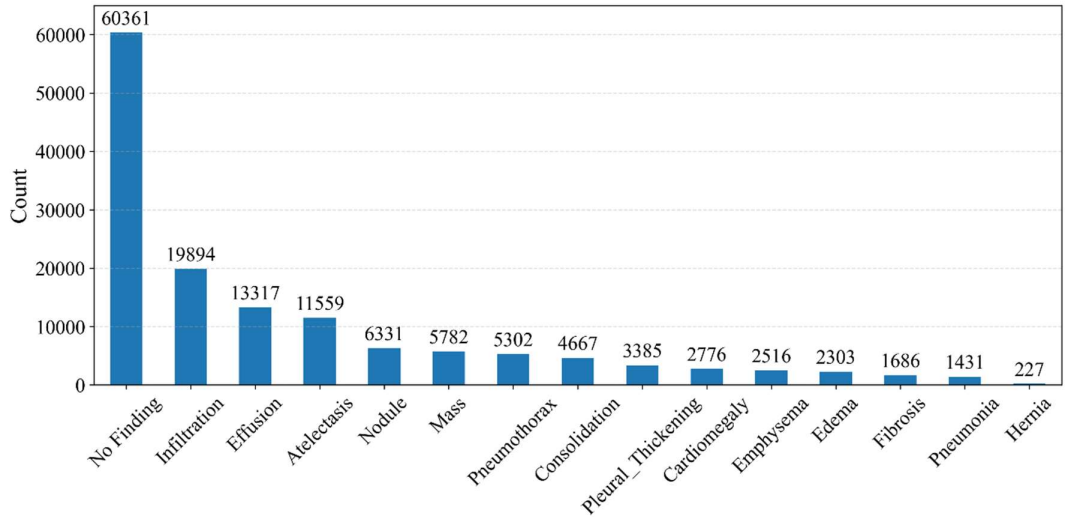

Figure 3. Unbalanced label distribution of the NIH CXR dataset.

To address class imbalance in the NIH dataset, we constructed a curated multi-label CXR subset derived from the NIH CXR database. Images were selected based on six target categories: five pulmonary abnormalities (Mass, Nodule, Pneumonia, Edema, and Fibrosis) and a normal category (No Finding). These conditions were selected because they are clinically important abnormalities commonly assessed on chest radiographs, ensuring that the classification task reflects meaningful diagnostic challenges. For each abnormality, we randomly sampled up to 2,000 images, using all available images when fewer than 2,000 were present, and we also randomly selected 2,000 No Finding images. After merging category-specific selections and removing duplicate images arising from multi-label overlap, the final dataset contained 10,486 unique chest X-ray images (Table 1). This sampling strategy reduces bias from highly prevalent labels and helps the model avoid overemphasizing common conditions. Notably, Mass and Nodule show strong co-occurrence in the matrix in Figure 4. This pattern is clinically plausible because both labels capture related radiographic manifestations of focal pulmonary lesions and may occur together in malignant processes.

Each image was encoded as a six-dimensional binary label vector in the fixed order [Mass, Nodule, Pneumonia, Edema, Fibrosis, No Finding]. Using the 70%, 15%, and 15% image level split, the final dataset contained 7,340 training images, 1,573 validation images, and 1,573 test images.

Table 1. Curated multi-label CXR subset derived from the NIH CXR database.

| Label | Number of Images |
| --- | --- |
| Mass | 2,375 |
| Nodule | 2,436 |
| Pneumonia | 1,431 |
| Edema | 2,052 |
| Fibrosis | 1,686 |
| No Finding | 2,000 |
| Total Unique Images | 10,486 |

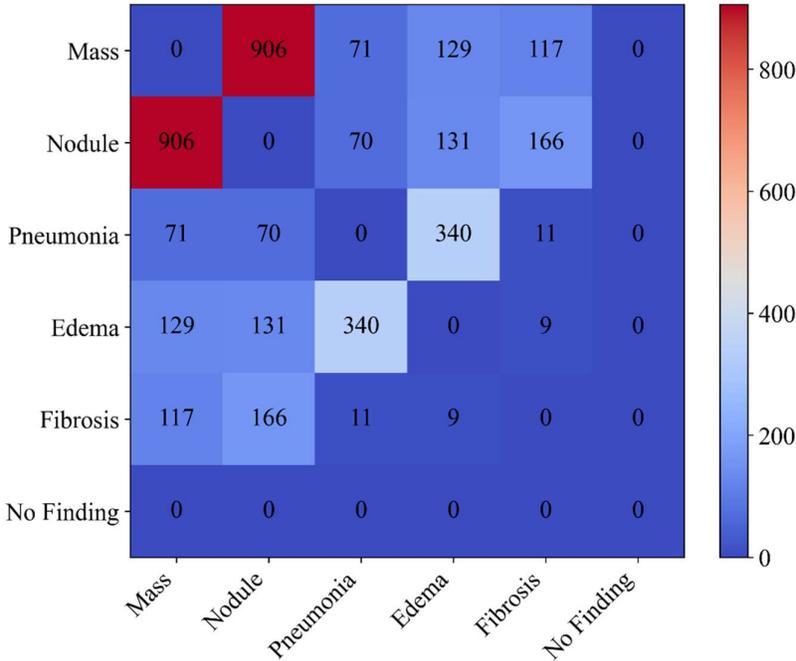

Figure 4. Co-occurrence heatmap of six labels in the curated NIH subset.

### 3.2. Lung segmentation using MedSAM

In our pipeline, each RSUA CXR image was first resized to a standardized resolution of 1024×1024 to match MedSAM's expected input resolution. The training images were then provided to MedSAM along with a coarse localization prompt. Specifically, we used bounding box prompts to specify the lung region, since boxes provide clearer spatial context than point prompts and reduce ambiguity when nearby structures have similar appearance. Data augmentation was limited to random perturbation of bounding box prompts, which encouraged robustness to imprecise localization. No additional geometric or photometric image augmentations were applied, as MedSAM's pretrained encoder and large-scale medical training corpus already provide strong invariance to appearance variability. During training, MedSAM was fine-tuned on the training set with performance monitored on the validation set. The fine-tuned model was then used to generate pixel-wise lung masks to suppress non-lung regions and extract lung-only images for subsequent classification. The segmentation accuracy of the fine-tuned MedSAM was evaluated using the test set. This approach allows efficient, reproducible lung delineation while retaining flexibility across variable acquisition conditions and patient anatomy.

### 3.3. Lung mask post-processing

The lung masks for the NIH CXR images were then generated using the fine-tuned MedSAM. Following [7], we expanded each mask to enlarge the region of interest (ROI). This ROI expansion helps preserve perihilar areas and adjacent thoracic anatomy, such as the cardiac silhouette, which can provide additional context for certain thoracic conditions. The enlargement was implemented using morphological operations, which can improve robustness and yield smooth mask boundaries [22, 23]. The process consisted of two steps:
1. Initial refinement: A morphological erosion with a disk structuring element with radius of 5 pixels was applied to suppress noise and remove small, isolated mask artifacts.
2. Controlled expansion: This was followed by a morphological dilation using two disk structuring elements with radii of 15 pixels and 50 pixels, respectively. This procedure produced two mask variants: a tight mask (M-15) and a loose mask (M-50), achieving controlled enlargement while maintaining smooth, continuous mask boundaries.

### 3.4. Classification models

We trained two deep learning neural networks, ResNet50 and DenseNet121, to evaluate multi-label abnormality classification on the original CXRs as well as two lung masked image sets. These architectures were selected because they are widely used, well-established backbones in prior CXR classification studies, providing reproducible baselines for comparison with related work. They also offer complementary strengths, since DenseNet's dense connectivity promotes feature reuse and can help capture subtle radiographic patterns, whereas ResNet's residual connections improve gradient flow and enhance training stability.

We formulated CXR diagnosis as a five-label classification problem focused on five clinically relevant abnormalities, including Mass, Nodule, Pneumonia, Edema, and Fibrosis, and each image could be associated with none, one or more of these labels. Under this setting, each CXR was represented by a binary label vector over the five targets, and the network produced a corresponding five-dimensional output in a single forward pass, with one logit per abnormality and sigmoid activation used to obtain label specific probabilities. The model was trained end-to-end with a multi-label objective that aggregates errors across labels, enabling shared feature learning across conditions while preserving independent predictions for each target. Compared with a per condition design in which separate binary classifiers are trained for each label, this single model approach is more computationally efficient for training and inference, reduces deployment complexity, and better reflects clinical scenarios where multiple findings can co-occur in the same radiograph. In addition, rather than treating "No Finding" as an independent label, we derived it at inference time as the complement of the predicted abnormalities, which helps avoid logically inconsistent outputs.

To quantify the effect of lung segmentation, we trained each backbone using three input variants: the original CXR, a tightly masked image (M-15), and a loosely masked image (M-50). Combining two backbones with three input variants yielded six classification settings. All models were initialized with ImageNet pretrained weights and fine-tuned end-to-end using binary cross entropy with logits (BCE) as the loss function. We optimized using Adam with an initial learning rate of $1 \times 10^{-4}$, a batch size of 4, and up to 30 epochs, with early stopping based on validation performance. To improve generalization, we applied training time data augmentation, including random horizontal flips, small rotations (±7 degrees), and mild intensity perturbations (up to 10%) to reflect variability in acquisition conditions. During validation, we automatically tuned label-specific decision thresholds to account for differences in prevalence and score calibration across abnormalities. These label-specific decision thresholds were selected using the precision–recall curve by choosing the threshold that maximized the F1 score for each abnormality. The selected thresholds were then fixed and applied to the corresponding test set for threshold-dependent evaluation. All experiments were run on a Dell workstation equipped with an NVIDIA GeForce RTX 4090 GPU (24 GB VRAM).

### 3.5. Evaluation metrics

Dice loss and binary cross entropy (BCE) loss are widely used to evaluate segmentation performance. The dice loss is defined as

$$L_{\text{Dice}} = 1 - \frac{2 |A \cap B|}{|A| + |B|} \quad (1)$$

where $L_{\text{Dice}}$ ranges from 0 to 1. A value of 0 indicates perfect overlap between the predicted mask and the ground truth, whereas 1 indicates no overlap. Here, $A$ denotes the binary mask predicted by the model and $B$ denotes the ground truth mask. BCE is defined as

$$L_{\text{BCE}} = -[y \log(p) + (1 - y) \log(1 - p)] \quad (2)$$

where $y \in \{0, 1\}$ is the ground truth binary label and $p$ is the predicted probability.

We evaluated CXR segmentation by averaging the Dice loss and BCE loss to leverage the complementary strengths of these two metrics [7]. This combined objective encourages accurate region overlap through Dice and calibrated pixel-level probabilities through BCE. Such a formulation is well suited to medical image segmentation, where capturing overall anatomy and localizing boundaries accurately are both clinically important. After mask expansion, we qualitatively assessed segmentation quality via visual inspection.

Classification performance was primarily evaluated using the area under the receiver operating characteristic curve (AUROC), because it measures discrimination across all possible decision thresholds and is widely reported in CXR diagnosis studies, enabling direct comparisons with prior work [3, 7]. AUROC was computed independently for each abnormality label on the test set of each split. We also report the macro AUROC, defined as the unweighted mean of AUROCs across the five abnormality categories, to avoid dominance by more prevalent labels and highlight performance on rarer or more challenging classes. For completeness, we additionally report macro averaged precision, recall, and F1 score to summarize overall model behavior.

Although No Finding was not explicitly trained, we derived a continuous No Finding score for each test image as:

$$\text{No Finding score} = 1 - \max_k (p_k) \quad (3)$$

where $k \in \{1,2,3,4,5\}$ indexes the five abnormalities, and $p_k$ denotes the predicted probability for abnormality $k$. This definition reflects the intuition that an image should be considered normal only when all abnormality probabilities are low. The resulting score was used to compute AUROC for No Finding. We did not adopt alternative formulations such as $1 - \prod(1 - p_k)$, which assume conditional independence among abnormalities, because the abnormality labels exhibit nontrivial co-occurrence patterns, as shown in Figure 4, making the independence assumption inappropriate.

To assess reliability, each experimental setting was repeated five times using five matched random 70:15:15 splits into training, validation, and test sets, and test performance is reported as mean ± standard deviation across runs. For statistical comparison between two model settings, we used a paired two-sided *t*-test on the split-level AUROC values with a significance level of α = 0.05. This test was applied to macro AUROC and No Finding AUROC comparisons reported in the Results.

## 4. Results

### 4.1. CXR segmentation

A total of 292 RSUA CXR image mask pairs were randomly split into 204 training, 44 validation, and 44 test pairs using a 70:15:15 ratio. MedSAM was fine-tuned with early stopping and training stopped at epoch 29 when validation performance plateaued. The best checkpoint was selected based on the minimum validation average loss, achieved at epoch 19. Using this checkpoint, the final average loss was 0.009 on the training set, 0.018 on the validation set, and 0.018 on the test set. The corresponding Dice and BCE losses were 0.008 and 0.010 for training, 0.017 and 0.019 for validation, and 0.016 and 0.020 for test, respectively. Figure 5 shows the training, validation, and test loss trajectories for the average loss, Dice loss, and BCE loss. Consistent with these curves, test losses remained higher than training losses throughout optimization, indicating a generalization gap between seen and held-out data.

We subsequently applied the fine-tuned MedSAM model to the curated set of 10,486 NIH CXR images (Table 1) to generate lung masks and masked CXR images for downstream classification. During this step, we identified a data consistency issue: a small subset of NIH CXR images stored in PNG format included an alpha transparency channel. To ensure a consistent input representation, we standardized these files by removing the alpha channel before inference. Qualitative inspection indicated that MedSAM produced anatomically plausible and consistent lung masks across a range of image qualities and acquisition conditions, as shown in Figure 6(a) and 6(b). The morphologically eroded mask, the two dilated masks (M-15 for tight masking and M-50 for loose masking), and the masked CXR image generated using the M-50 are shown in Figure 6(c)–6(f).

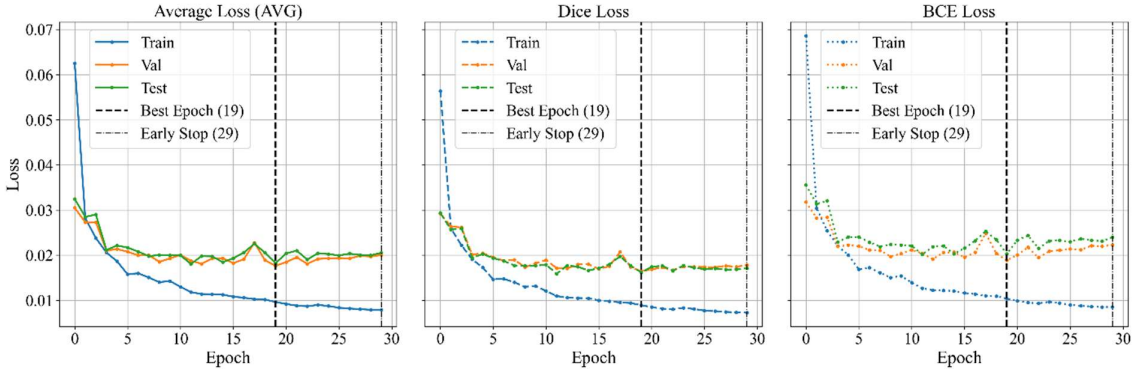

Figure 5. Training, validation, and test loss curves for CXR lung segmentation, including average loss, Dice loss, and BCE loss. The best checkpoint was selected at epoch 19 and achieved a test average loss of 0.018.

### 4.2. Classification

We evaluated six classification settings defined by two backbone architectures, DenseNet121 and ResNet50, and three input variants: the original images, tightly masked images (M-15), and loosely masked images (M-50). Table 2 summarizes classification performance across the five matched 70:15:15 splits, reported as mean ± standard deviation. Abnormality level performance is reported using AUROC values computed independently for each abnormality on the test set of each split, along with the macro AUROC across the five abnormality labels. Among all configurations, ResNet50 trained on the original images achieved the highest macro AUROC of 79.4 ± 0.4. Tight masking (M-15) consistently reduced macro AUROC for both backbones, decreasing DenseNet121 from 78.3 ± 0.6 to 77.7 ± 0.6 and ResNet50 from 79.4 ± 0.4 to 78.6 ± 0.8, suggesting that overly aggressive spatial restriction can remove diagnostically relevant contextual cues. Loose masking partially recovered performance relative to M-15, with DenseNet121 (M-50) reaching 78.2 ± 0.5 and ResNet50 (M-50) reaching 79.2 ± 0.9, supporting the interpretation that preserving additional perihilar and peripheral context mitigates the information loss introduced by tight masking. Importantly, ResNet50 and ResNet50 (M-50) did not differ significantly in macro AUROC based on a two-sided paired *t*-test, indicating that expanded masking does not materially change overall abnormality discrimination for this backbone.

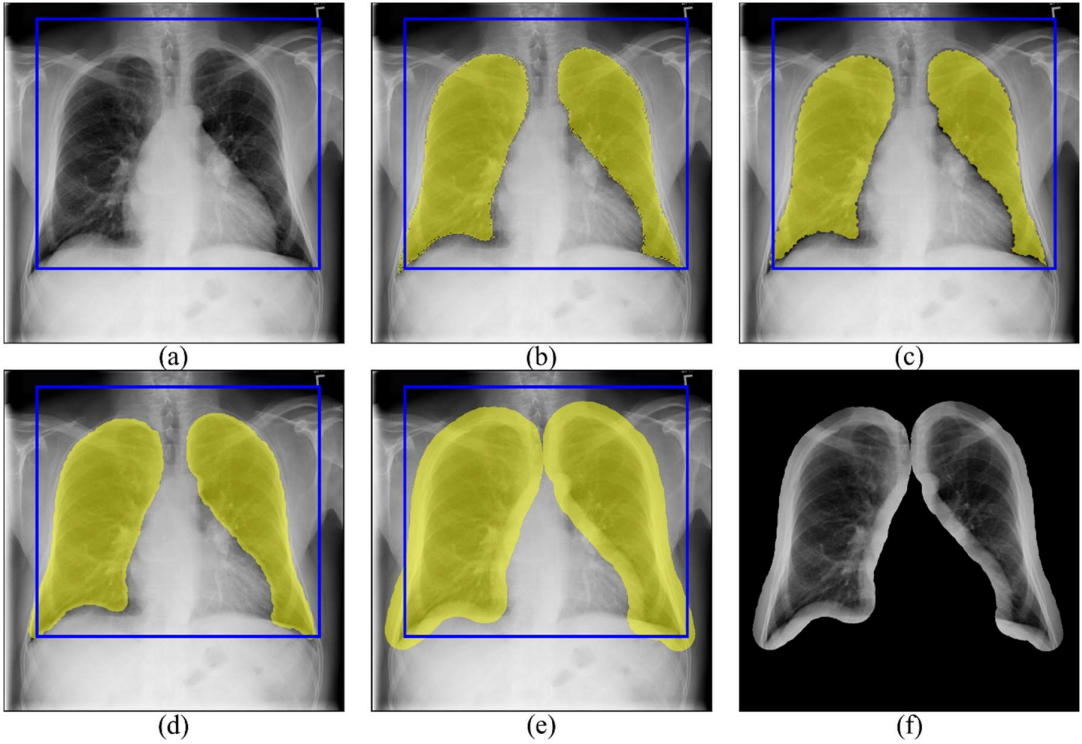

Figure 6. Qualitative example of CXR lung segmentation and morphological mask expansion: (a) original CXR with prompt box, (b) raw MedSAM lung mask overlay, (c) eroded mask overlay (disk radius = 5 pixels), (d) dilated mask overlay (M-15, tight masking, disk radius 15 pixels), (e) dilated mask overlay (M-50, loose masking, disk radius 50 pixels), and (f) masked CXR image generated using M-50.

Table 2. Abnormality-level and No Finding AUROC (×100), training time, and an external baseline reference for all six approaches. M-15 and M-50 denote tight and loose masking, respectively. FT ResNet-50 (2019) is included for reference, and results are not directly comparable due to differences in experimental settings and evaluation protocols [3].

| Label | DenseNet 121 | DenseNet 121 (M-15) | DenseNet 121 (M-50) | ResNet50 | ResNet50 (M-15) | ResNet50 (M-50) | FT ResNet-50 (2019) |
|---|---|---|---|---|---|---|---|
| Mass | 80.1±1.1 | 80.1±0.7 | 81.1±0.6 | 81.1±0.6 | 80.9±1.1 | 81.2±0.7 | 82.2±1.0 |
| Nodule | 71.8±2.6 | 72.0±1.2 | 72.1±0.9 | 72.4±1.3 | 72.4±1.3 | 72.4±0.9 | 72.6±0.9 |
| Pneumonia | 70.6±0.7 | 68.2±0.9 | 69.8±1.5 | 71.1±1.1 | 69.4±1.0 | 70.9±0.6 | 74.4±1.6 |
| Edema | 90.9±0.1 | 90.2±0.2 | 90.0±0.6 | 91.6±0.1 | 90.6±0.4 | 91.2±0.5 | 89.1±0.4 |
| Fibrosis | 78.1±1.0 | 78.1±0.9 | 78.2±0.2 | 80.9±0.6 | 79.8±1.0 | 80.2±1.2 | 80.0±0.9 |
| Macro AUROC | 78.3±0.6 | 77.7±0.6 | 78.2±0.5 | **79.4±0.4** | 78.6±0.8 | **79.2±0.9** | 79.7±0.5 |
| No Finding | 77.7±1.4 | 77.7±1.1 | 77.3±1.4 | 77.2±1.3 | 76.9±1.1 | **79.1±0.6** | 76.9±0.5 |
| Training time (30 epochs) | ~65.3 min | ~63.0 min | ~58.0 min | ~54.0 min | ~52.0 min | ~50.0 min | - |

Across individual abnormality categories, Edema consistently achieved the highest AUROC across settings, peaking at 91.6 ± 0.1 with ResNet50 trained on the original images, consistent with its relatively distinctive radiographic appearance. In contrast, Pneumonia and Nodule yielded lower AUROC values across settings, reflecting their visual heterogeneity and potential label ambiguity in chest radiographs. For comparison with prior work, [3] reported a fine-tuned ResNet50 baseline with a macro AUROC of 79.7 ± 0.5, which is comparable to the best setting in our study given differences in experimental design and evaluation protocols.

Because No Finding was derived rather than explicitly trained, we evaluated it separately from the five abnormalities. As shown in Table 2, ResNet50 (M-50) achieved the highest No Finding AUROC of 79.1 ± 0.6. This value is higher than the ResNet50 baseline reported in [3], although the two results are not directly comparable due to differences in experimental settings. Interestingly, compared with ResNet50 trained on the original images, ResNet50 (M-50) achieved a statistically significantly higher No Finding AUROC based on a two-sided paired $t$-test, even though the macro abnormality AUROC for the two settings was comparable. This result suggests that loose masking improves discrimination of normal cases and highlights a trade-off between abnormality-specific classification and normal versus abnormal screening performance.

To complement AUROC, we report macro averaged precision, recall, and F1 score across the five abnormalities in Table 3 to characterize threshold-dependent performance. Precision measures the proportion of predicted positives that are correct, recall measures the proportion of true positives that are recovered, and F1 summarizes the balance between them. Consistent with the AUROC analysis, ResNet50 trained on the original images achieved the highest macro F1 score of 0.539 ± 0.008. Tight masking reduced macro F1 for both backbones, whereas loose masking partially recovered performance. In contrast, ResNet50 (M-50) achieved the highest No Finding F1, highlighting the possible task-dependent benefit of loose masking for normal case screening, although the improvement was modest and not statistically significant across the five matched splits. It is likely because F1 is threshold-dependent and reflects performance at a single operating point, it can be less sensitive than AUROC to small but consistent shifts in ranking-based discrimination.

Table 3. Threshold-dependent classification performance shown as mean ± standard deviation across five matched splits, reporting macro Precision (P), Recall (R), and F1 for five abnormalities and No Finding Precision (P), Recall (R), and F1.

| Metric | DenseNet 121 | DenseNet 121 (M-15) | DenseNet 121 (M-50) | ResNet50 | ResNet50 (M-15) | ResNet50 (M-50) |
|---|---|---|---|---|---|---|
| Macro P | 0.445±0.011 | 0.450±0.012 | 0.452±0.010 | 0.471±0.006 | 0.442±0.005 | 0.454±0.011 |
| Macro R | 0.612±0.008 | 0.580±0.023 | 0.616±0.009 | 0.632±0.012 | 0.596±0.013 | 0.628±0.025 |
| Macro F1 | 0.517±0.006 | 0.506±0.003 | 0.521±0.006 | **0.539±0.008** | 0.508±0.007 | 0.527±0.004 |
| No Finding P | 0.512±0.025 | 0.516±0.028 | 0.524±0.034 | 0.526±0.034 | 0.510±0.029 | 0.526±0.033 |
| No Finding R | 0.618±0.018 | 0.612±0.047 | 0.621±0.055 | 0.613±0.054 | 0.642±0.040 | 0.633±0.042 |
| No Finding F1 | 0.560±0.010 | 0.560±0.005 | 0.569±0.012 | **0.566±0.011** | 0.569±0.007 | **0.574±0.008** |

Figure 7 provides a direct visual comparison of ROC behavior between ResNet50 trained on the original images and ResNet50 with loose masking (M-50), clarifying the metric level trade-offs observed in Tables 2. Across the five abnormalities, the ROC curves largely overlap between the two settings, indicating comparable abnormality-level discrimination and reinforcing that macro AUROC difference between ResNet50 and ResNet50 (M-50) is not statistically significant. In contrast, the derived No Finding ROC curve (dashed line) shifts upward under M-50, yielding a visibly higher AUROC than the original image model. This improvement reflects lower false positive rates at comparable true positive rates and visually confirms that loose masking enhances normal case discrimination.

Training efficiency is also summarized in Table 2. Over 30 epochs, training time ranged from approximately 50 to 66 minutes across settings, with ResNet50 consistently faster than DenseNet121 under comparable conditions. Masking reduced training time for both architectures, and the shortest runtime was observed for ResNet50 (M-50), suggesting that lung focused preprocessing can lower computational cost.

Finally, the predicted label co-occurrence pattern in the test set for ResNet50 (M-50) is shown in Figure 8 and is broadly consistent with the overall dataset co-occurrence structure in Figure 4. Similar patterns were observed for the other settings, suggesting that the models generally reflect the dataset's multi-label structure in their predictions. To quantify this agreement, we compared the original dataset and predicted test set co-occurrence matrices (Figure 4 and Figure 8) using a Mantel test based on Spearman correlation of the off-diagonal entries with permutation testing. The predicted co-occurrence matrix on the test set showed a strong and statistically significant correlation with the original dataset structure, with Spearman ρ = 0.89 and p = 0.005.

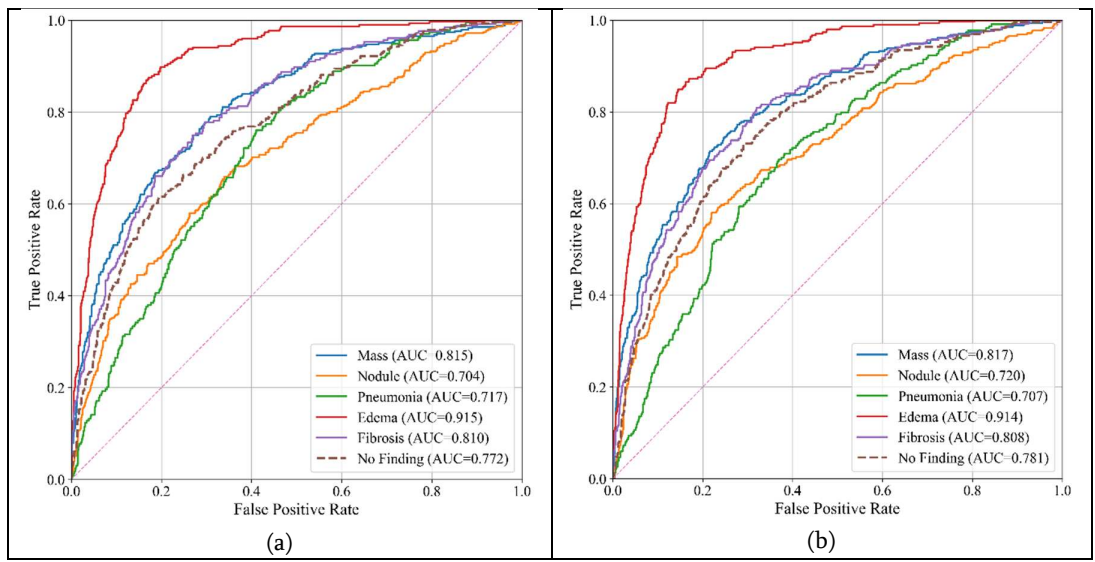

Figure 7. ROC curves for the five abnormalities, along with the derived No Finding ROC curve (dashed line): (a) ResNet50 trained on the original images; (b) ResNet50 (M-50).

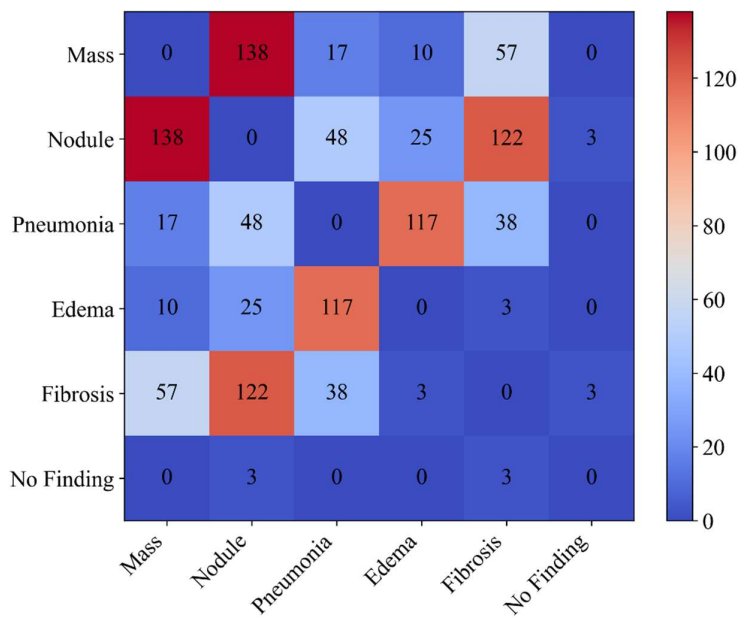

Figure 8. Co-occurrence heatmap of six predicted conditions on the test set for ResNet50 (M-50).

# 5. Discussion

This study systematically examined how network architecture interacts with lung focused preprocessing in multi-label CXR classification, using MedSAM as a foundation model to generate lung masks and regulate the amount of retained anatomical context. Overall, the segmentation module provides a principled mechanism to constrain classification to anatomically relevant regions and reduce reliance on spurious background cues. In practice, masking also improved computational efficiency by limiting irrelevant variability, as reflected by consistently shorter training times when masked inputs were used. Our findings support the broader view that foundation model-based segmentation can serve as a robust and reproducible preprocessing component for CXR analysis, particularly when interpretability and regional relevance are priorities.

However, the classification results show that segmentation is not a universally beneficial preprocessing step, and its effect depends on both the backbone and the clinical objective. Across all six settings, architecture choice remained the dominant factor for abnormality level discrimination. ResNet50 trained on the original images achieved the highest macro AUROC and macro F1, indicating that residual connections can effectively leverage global contextual cues that may be partially attenuated by masking. With tight masking, performance degraded consistently for both backbones, suggesting that overly aggressive spatial restriction removes diagnostically relevant information, including perihilar, pleural, and peripheral context that can contribute to abnormality recognition. Expanded masking mitigated this effect by reintroducing a controlled margin of surrounding anatomy. Importantly, ResNet50 (M-50) and ResNet50 did not differ significantly in macro AUROC, reinforcing that loose masking does not significantly reduce overall abnormality discrimination for this backbone.

The derived No Finding analysis revealed a clinically meaningful trade-off that is not captured by macro abnormality AUROC alone. Although ResNet50 and ResNet50 (M-50) achieved comparable macro AUROC, loose lung masking produced a significantly higher No Finding AUROC, indicating improved separation of normal and abnormal cases. Together, these results suggest that lung segmentation can redistribute model behavior by suppressing non-lung background cues, which benefits screening-oriented tasks that prioritize normal case identification but does not necessarily enhance fine grained, abnormality-specific discrimination. In practical terms, the optimal input strategy depends on whether the intended application emphasizes abnormality-specific discrimination, or normal case screening. This interpretation is consistent with prior COVID-19 era pipelines reporting that segmentation modules can improve classification performance for normal, COVID-19, and pneumonia categories [7].

Abnormality-specific patterns also align with known radiographic characteristics and dataset limitations. Edema consistently achieved the highest AUROC across approaches, reflecting relatively distinctive imaging cues compared with the other target labels. In contrast, Pneumonia and Nodule produced lower AUROC values across settings, consistent with their visual heterogeneity and the ambiguity of the NIH label schema. More broadly, several factors likely contributed to the moderate performance ceiling observed in this study. First, NIH labels are known to be imperfect, and prior work has reported label extraction performance around 0.90 on external validation data [4], which may effectively limit achievable discrimination because even an ideal model cannot reliably learn from inconsistent ground truth. Second, multi-label classification is inherently more challenging than single label prediction because findings can co-occur, share overlapping radiographic cues, and exhibit correlated label patterns, which increases learning difficulty and evaluation uncertainty. Third, several NIH labels represent broad clinical concepts rather than tightly defined entities, with Pneumonia as a common example. Radiographic pneumonia can reflect heterogeneous etiologies and patterns, can overlap with edema or atelectasis, and often depends on clinical context not available from the image alone, all of which increases label noise and weakens alignment between visual evidence and annotations.

Several methodological limitations should be considered when interpreting the segmentation component and its downstream effects. Although MedSAM produced anatomically plausible lung masks in most cases, segmentation fidelity may degrade under severe image noise or when pathology obscures lung boundaries, motivating evaluation across a broader range of image qualities and multi-institutional data. In addition, although masking reduced training time, practical model development remained constrained by GPU resources, requiring trade-offs among batch size, model complexity, and training stability. These operational factors are relevant for reproducibility and for scaling the pipeline to larger datasets or more complex architectures.

Taken together, our results suggest that lung masking should not be applied as a default preprocessing step but rather treated as a controllable spatial prior whose value depends on the backbone and the clinical goal. For abnormality-specific classification, ResNet50 trained on original images remained the most effective setting in our experiments. For screening-oriented tasks that prioritize normal versus abnormal separation, loose lung masking provides a measurable advantage by improving No Finding discrimination while maintaining comparable macro abnormality AUROC.

A limitation of this study is that the NIH CXR dataset was split at the image level rather than at the patient level. Because multiple radiographs from the same patient may be present in the dataset, this approach introduces the possibility of patient overlap between training and test sets, which could lead to optimistic estimates of absolute performance. However, our primary objective was to compare relative performance trends across masking strategies and backbone architectures under matched experimental conditions. All model variants were trained and evaluated using identical splits, and conclusions regarding the task-dependent and architecture-dependent effects of lung masking are therefore expected to remain valid. Future work will repeat the full evaluation under patient-level splits to confirm that the observed masking trends persist when patient overlap is eliminated.

# 6. Conclusion

This study developed and evaluated a segmentation guided CXR classification pipeline that integrates MedSAM as a foundation model for lung segmentation and region focused preprocessing. Although MedSAM was fine-tuned and evaluated on a relatively small RSUA CXR dataset, segmentation performance was reliable and produced anatomically plausible lung masks that generalized across diverse image appearances. Through systematic experiments varying backbone architecture and mask tightness, we showed that segmentation-based preprocessing is not universally optimal, but instead functions best as a controllable spatial prior whose value depends on both the model and the clinical objective.

Across all configurations, ResNet50 trained on original images achieved the strongest overall abnormality-level performance, indicating that preserving full image context benefits abnormality-specific discrimination for this backbone. Tight masking consistently degraded performance, suggesting that overly aggressive spatial restriction removes diagnostically relevant contextual cues. Loose masking mitigated this degradation and improved training efficiency. It also significantly improved No Finding discrimination for screening, while maintaining comparable macro abnormality AUROC. These results highlight a practical trade-off between abnormality-specific classification and normal case screening performance and emphasize that masking strategy should be selected to match the intended use case rather than applied uniformly.

Future work will further examine these effects under patient-level splits to fully eliminate potential data leakage and to validate the robustness of the observed trends. In addition, incorporating patient-level metadata such as age and sex, expanding the label set, and refining abnormality definitions may further improve both model performance and clinical utility. Additional studies should also evaluate alternative architectures to determine whether the observed masking trade-offs persist under different representation learning paradigms.